*Research Note*

# Applying GSAT to Non-Clausal Formulas

**Roberto Sebastiani** RSEBA@DIST.UNIGE.IT
*Mechanized Reasoning Group, DIST, viale Causa 13, 16145 Genova, Italy.*
*Mechanized Reasoning Group, IRST, loc. Pantè, 38050 Povo, Trento, Italy.*

## Abstract

In this paper we describe how to modify GSAT so that it can be applied to non-clausal formulas. The idea is to use a particular "score" function which gives the number of clauses of the CNF conversion of a formula which are false under a given truth assignment. Its value is computed in linear time, without constructing the CNF conversion itself. The proposed methodology applies to most of the variants of GSAT proposed so far.

## 1. Introduction

GSAT (Selman, Levesque, & Mitchell, 1992; Selman & Kautz, 1993) is an incomplete model-finding algorithm for clausal propositional formulas which performs a randomized local search. GSAT has been shown to solve many "hard" problems much more efficiently than other traditional algorithms like, e.g., DP (Davis & Putnam, 1960). Since GSAT applies only to clausal formulas, using it to find models for ordinary propositional formulas requires some previous clausal-form conversion. This requires extra computation (which can be extremely heavy if the "standard" clausal conversion is used). Much worse, clausal-form conversion causes either a large increase in the size of the input formula or an enlargement of the search space.

In this paper we describe how to modify GSAT so that it can be applied to non-clausal formulas *directly*, i.e., with no previous clausal form conversion. An extended version of the paper (Sebastiani, 1994) provides the proofs of the theorems and a detailed description of the algorithm introduced.

This achievement could enlarge GSAT's application domain. Selman et al. (1992) suggest that some traditional AI problems can be formulated as model-finding tasks; e.g., visual interpretation (Reiter & Mackworth, 1989), planning (Kautz & Selman, 1992), generation of "vivid" knowledge representation (Levesque, 1986). It is often the case that non-clausal representations are more compact for such problems. For instance, each rule in the form "$\bigwedge_i \phi_i \supset \psi$" gives rise to several distinct clauses if some $\phi_i$ are disjuncts or $\psi$ is a conjunct. In automated theorem proving (a.t.p.) some applications of model-finding have been proposed (see, e.g., (Artosi & Governatori, 1994; Klingerbeck, 1994)). For instance, some decision procedures for decidable subclasses of first-order logic iteratively perform non-clausal model-finding for propositional instances of the input formulas (Jeroslow, 1988). More generally, some model-guided techniques for proof search, like *goal deletion* (Ballantyne & Bledsoe, 1982), *false preference*, or *semantic resolution* (Slaney, 1993), seem to be applicable to non-clausal a.t.p. as well.

```
procedure GSAT(φ)
    for j := 1 to Max-tries do
        T := initial(φ)
        for k := 1 to Max-flips do
            if T ⊨ φ
            then return T
            else Poss-flips := hill-climb(φ, T)
                 V := pick(Poss-flips)
                 T := flip(V,T)
                 UpdateScores(φ, V)
        end
    end
    return "no satisfying assignment found".
```

Figure 1: A general schema for GSAT.

## 2. GSAT

If $\phi$ is a clausal propositional formula and $T$ is a truth assignment for the variables of $\phi$, then the number of clauses of $\phi$ which are falsified by $T$ is called the *score* of $T$ for $\phi$ ($score(T, \phi)$). $T$ satisfies $\phi$ iff $score(T, \phi) = 0$. The notion of score plays a key role in GSAT, as it is considered as the "distance" from a truth assignment to a satisfying one.

The schema of Figure 2 describes GSAT as well as many of its possible variants. We use the notation from (Gent & Walsh, 1993). GSAT performs an iterative search for a satisfying truth assignment for $\phi$, starting from a random assignment provided by *initial()*. At each step, the successive assignment is obtained by flipping (inverting) the truth value of one single variable $V$ in $T$. $V$ is chosen to minimize the score. Let $T_i$ be the assignment obtained from $T$ by flipping its $i$-th variable $V_i$. *hill-climb()* returns the set *Poss-flips* of the variables $V_r$ which minimize $score(T_r, \phi)$. If the current values of $\Delta s_i = score(T_i, \phi) - score(T, \phi)$ are stored for every variable $V_i$, then *hill-climb()* simply returns the set of the variables $V_r$ with the best $\Delta s_r$. *pick()* chooses randomly one of such variables. *flip()* returns $T$ with $V$'s value flipped. After each flipping, *UpdateScores()* updates the values of $\Delta s_i$, for all $i$.

This paper exploits the observation that the functions *initial()*, *hill-climb()*, *pick()* and *flip()* do not depend on the structure of the input formula $\phi$, and that the computation of the scores is the only step where the input formula $\phi$ is required to be in clausal form. The idea is thus to find a suitable notion of score for non-clausal formulas, and an efficient algorithm computing it.

## 3. An extended notion of score

Let *cnf($\varphi$)* be the result of converting a propositional formula $\varphi$ into clausal form by the standard method (i.e., by applying the rules of De Morgan). Then the following definition extends the notion of score to all propositional formulas.

**Definition 3.1** *The* score *of a truth assignment $T$ for a propositional formula $\varphi$ is the number of the clauses of cnf($\varphi$) which are falsified by $T$.*

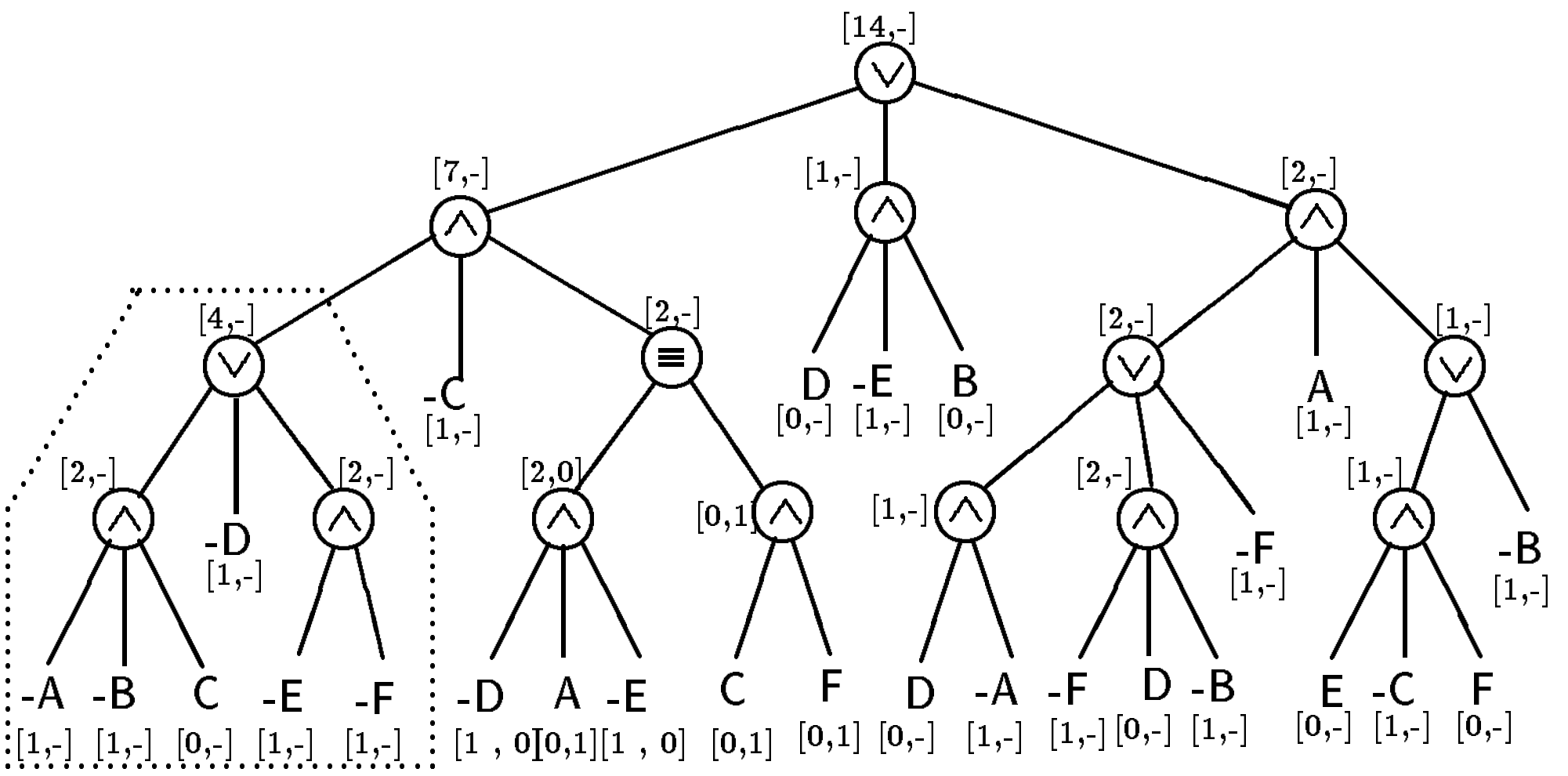

Figure 2: The computation tree of $s(T, \varphi)$.

*cnf()* represents the "natural" clausal form conversion. *cnf($\varphi$)* has the same number of propositional variables as $\varphi$ and it is logically equivalent to $\varphi$. The problem with *cnf()* is the exponential size growth of *cnf($\varphi$)*, that is, $|cnf(\varphi)| = O(2^{|\varphi|})$. Definition 3.1 overcomes such a problem, for it is possible to introduce a linear-time computable function $s(T, \varphi)$ which gives the score of $T$ for a formula $\varphi$. This is done *directly*, i.e., without converting $\varphi$ into clausal form. We define $s(T, \varphi)$ recursively as follows: [1]

| $\varphi$ | $s(T, \varphi)$ | $s^-(T, \varphi)$ |
|---|---|---|
| $\varphi$ *literal* | $\begin{cases} 0 & if\ T \models \varphi \\ 1 & otherwise \end{cases}$ | $\begin{cases} 1 & if\ T \models \varphi \\ 0 & otherwise \end{cases}$ |
| $\neg\varphi_1$ | $s^-(T, \varphi_1)$ | $s(T, \varphi_1)$ |
| $\bigwedge_k \varphi_k$ | $\sum_k s(T, \varphi_k)$ | $\prod_k s^-(T, \varphi_k)$ |
| $\bigvee_k \varphi_k$ | $\prod_k s(T, \varphi_k)$ | $\sum_k s^-(T, \varphi_k)$ |
| $\varphi_1 \supset \varphi_2$ | $s^-(T, \varphi_1) \cdot s(T, \varphi_2)$ | $s(T, \varphi_1) + s^-(T, \varphi_2)$ |
| $\varphi_1 \equiv \varphi_2$ | $s^-(T, \varphi_1) \cdot s(T, \varphi_2) +$ $s(T, \varphi_1) \cdot s^-(T, \varphi_2)$ | $(s(T, \varphi_1) + s^-(T, \varphi_2)) \cdot$ $(s^-(T, \varphi_1) + s(T, \varphi_2))$ |

$s^-(T, \varphi_k)$ is $s(T, \neg\varphi_k)$. The distinction between $s(T, \varphi_k)$ and $s^-(T, \varphi_k)$ is due to the *polarity* of the current subformula $\varphi_k$. During the computation of $s(T, \varphi)$, a call to the function $s(T, \varphi_j)$ [$s^-(T, \varphi_j)$] is invoked iff $\varphi_j$ is a positive [negative] subformula of $\varphi$.

**Example 3.1** Figure 2 represents the computation tree of the score of a truth assignment $T$ for the formula $\varphi$ :

$$(((\neg A \wedge \neg B \wedge C) \vee \neg D \vee (\neg E \wedge \neg F)) \wedge \neg C \wedge ((\neg D \wedge A \wedge \neg E) \equiv (C \wedge F))) \vee$$
$$(D \wedge \neg E \wedge B) \vee (((D \wedge \neg A) \vee (\neg F \wedge D \wedge \neg B) \vee \neg F) \wedge A \wedge ((E \wedge \neg C \wedge F) \vee \neg B)).$$

$T$ assigns "true" to all the variables of $\varphi$. The information in square brackets associated to any subformula $\varphi_j$ represents $[s(T, \varphi_j), s^-(T, \varphi_j)]$. For instance, if we consider the small subtree in the left of Figure 2, then the score is computed in the following way:

1. Notice that the definition of $s(T, \varphi)$ can be easily extended to formulas involving other connectives (e.g., *nand*, *nor*, *xor*, *if-then-else*, ...) or more complicate boolean functions.

$$
\begin{array}{ll}
s(T,\ (\neg A \wedge \neg B \wedge C) \vee \neg D \vee (\neg E \wedge \neg F)\ ) = & ;\ s(T, \bigvee_k \varphi_k) = \prod_k s(T, \varphi_k) \\
s(T, \neg A \wedge \neg B \wedge C) \cdot s(T, \neg D) \cdot s(T, \neg E \wedge \neg F) = & ;\ s(T, \bigwedge_k \varphi_k) = \sum_k s(T, \varphi_k) \\
(s(T, \neg A) + s(T, \neg B) + s(T, C)) \cdot s(T, \neg D) \cdot (s(T, \neg E) + s(T, \neg F)) = & ;\ \textit{literals} \\
(1 + 1 + 0) \cdot 1 \cdot (1 + 1) = 4. &
\end{array}
$$

Notice that $cnf(\varphi)$ is 360 clauses long. □

**Theorem 3.1** *Let $\varphi$ be a propositional formula and $T$ a truth assignment for the variables of $\varphi$. Then the function $s(T, \varphi)$ gives the score of $T$ for $\varphi$.*

The proof follows from the consideration that, for any truth assignment $T$, the set of the false clauses of *cnf*($\varphi_1 \vee \varphi_2$) is the cross product between the two sets of the false clauses of *cnf*($\varphi_1$) and *cnf*($\varphi_2$).

**Theorem 3.2** *Let $\varphi$ be a propositional formula and $T$ a truth assignment for the variables of $\varphi$. Then the number of operations required for calculating $s(T, \varphi)$ grows linearly with the size of $\varphi$.*

The proof follows from the fact that, if $Time(s^{\pm}(\varphi_i, T))$ is the number of operations required for computing both $s(T, \varphi_i)$ and $s^{-}(T, \varphi_i)$, and if $Time(s^{\pm}(\varphi_i, T)) \leq a_i \cdot |\varphi_i| + b_i$, then $Time(s^{\pm}(\varphi_1 \diamond \varphi_2, T)) \leq max_i(a_i) \cdot |\varphi_1 \diamond \varphi_2| + 2 \cdot max_i(b_i) + 6$, for any $\diamond \in \{\wedge, \vee, \supset, \equiv\}$.

The number of operations required for computing the score of an assignment $T$ for a clausal formula $\phi$ is $O(|\phi|)$. If $\phi = cnf(\varphi)$, then $|\phi| = O(2^{|\varphi|})$. Thus the standard computation of the score of $T$ for $\phi$ requires $O(2^{|\varphi|})$ operations, while $s(T, \varphi)$ performs the same result directly in linear time.

## 4. GSAT for non-clausal formulas

It follows from Sections 2, 3 that we can extend GSAT to non-clausal formulas $\varphi$ by simply using the extended notion of score of Definition 3.1. Let NC-GSAT (non-clausal GSAT) be a new version of GSAT in which the scores are computed by some implementation of the function *s()*. Then it follows from Theorem 3.1 that in *NC-GSAT*($\varphi$) the function *hill-climb()* always returns the same sets of variables as in *GSAT*(*cnf*($\varphi$)), so that *NC-GSAT*($\varphi$) performs the same flips and returns the same result as *GSAT*(*cnf*($\varphi$)). Theorem 3.2 ensures that every score computation is performed in linear time.

The current implementation of GSAT (Selman & Kautz, 1993) provides a highly-optimized implementation of *Updatescores*($\phi, V$), which analyzes only the clauses which the last-flipped variable $V$ occurs in. This allows a strong reduction in computational cost. In (Sebastiani, 1994) we describe in detail an analogous optimized version of the updating procedure for NC-GSAT, called *NC-Updatescores*($\varphi, V$), and prove the following properties:

(i) if $\varphi$ is in clausal form, i.e., $\varphi = cnf(\varphi)$, then *NC-UpdateScores*($\varphi, V$) has the same complexity as *UpdateScores*($\varphi, V$);

(ii) if $\phi = cnf(\varphi)$, then *NC-UpdateScores*($\varphi, V$) is $O(|\varphi|)$. *UpdateScores*($\phi, V$) is $O(2^{|\varphi|})$.

The latter mirrors the complexity issues presented in Section 3.

The idea introduced in this paper can be applied to most variants of GSAT. In "CSAT" (Cautious SAT) *hill-climb()* returns all the variables which cause a decrease of the score; in "DSAT" (Deterministic SAT) the function *pick()* performs a deterministic choice; in "RSAT" (Random walk SAT) the variable is picked randomly among *all* the variables; in "MSAT" (Memory SAT) *pick()* remembers the last flipped variable and avoids picking it. All these variants, proposed in (Gent & Walsh, 1992, 1993), can be transposed into NC-GSAT as well, as they are independent of the structure of the input formula. Selman and Kautz (1993) suggest some variants which improve the performance and overcome some problems, such as that of escaping local minima. The strategy "*Averaging in*" suggests a different implementation of the function *initial()*: instead of a random assignment, *initial()* returns a bitwise average of the best assignments of the two latest cycles. This is independent of the form of the input formula. In the strategy "*random walk*" the sequence *hill-climb()* - *pick()* is substituted with probability $p$ by a simpler choice function: "choose randomly a variable occurring in some unsatisfied clause". This idea can be transposed into NC-GSAT as well: "choose randomly a branch passing only for nodes whose score is different from zero, and pick the variable at the leaf".

One final observation is worth making. In order to overcome the exponential growth of CNF formulas, some algorithms have been proposed (Plaisted & Greenbaum, 1986; de la Tour, 1990) which convert propositional formulas $\varphi$ into polynomial-size clausal formulas $\psi$. Such methods are based on the introduction of new variables, each representing a subformula of the original input $\varphi$. Unfortunately, the issue of size-polynomiality is valid only if no "$\equiv$" occurs in $\varphi$, as the number of clauses of $\psi$ grows exponentially with the number of "$\equiv$" in $\varphi$. Even worse, the introduction of $k$ new variables enlarges the search space by a $2^k$ factor and reduces strongly the solution ratio. In fact, any model for $\psi$ is also a model for $\varphi$, but for any model of $\varphi$ we only know that *one* of its $2^k$ extensions is a model of $\psi$ (Plaisted & Greenbaum, 1986).

## Acknowledgements

Fausto Giunchiglia and Enrico Giunchiglia have given substantial and continuous feedback during the whole development of this paper. Toby Walsh provided important feedback about a previous version of this paper. Aaron Noble, Paolo Pecchiari, and Luciano Serafini helped with the final revision. Bart Selman and Henry Kautz are thanked for assistance with the GSAT code.